\documentclass[journal]{IEEEtran}
\ifCLASSINFOpdf
\else
\fi

\usepackage{subfigure}
\usepackage{epsfig}
\usepackage{graphicx}
\usepackage{amsmath}
\usepackage{amssymb}
\usepackage{algorithm, algorithmic}
\graphicspath{{figure/}}   
\usepackage{multirow}
\newcommand{\tabincell}[2]{\begin{tabular}{@{}#1@{}}#2\end{tabular}}
\usepackage{url}
\begin{document}
%
\title{	A CNN with Noise Inclined Module and Denoise Framework for Hyperspectral Image Classification}
%
%
%

\author{Zhiqiang~Gong,
				Ping~Zhong,~\IEEEmembership{Senior Member,~IEEE},
				Jiahao~Qi,       
				and~Panhe~Hu
\thanks{Manuscript received XX, 2022; revised XX, 2022. This work was supported by the Natural Science Foundation of China under Grant 62001502.(Corresponding author: Panhe Hu)}
\thanks{ Z. Gong is with the National Innovation Institute of Defense Technology,
Chinese Academy of Military Science, Beijing 100000, China. e-mail: (gongzhiqiang13@nudt.edu.cn).}
\thanks{P. Zhong, J. Qi, and P. Hu are with  the National Key Laboratory of Science and Technology on ATR, College of Electrical Science and Technology, National University of Defense Technology, Changsha 410073, China (e-mail: zhongping@nudt.edu.cn; hupanhe13@nudt.edu.cn).}
}

%
%

\markboth{IEEE LATEX,~Vol X, 2022}%
{Shell \MakeLowercase{\textit{et al.}}: Bare Demo of IEEEtran.cls for IEEE Journals}
%



\maketitle

\begin{abstract}
Deep Neural Networks have been successfully applied in hyperspectral image classification. However, most of prior works adopt general deep architectures while ignore the intrinsic structure of the hyperspectral image, such as the physical noise generation.
This would make these deep models unable to generate discriminative features and provide impressive classification performance.
To leverage such intrinsic information, this work develops a novel deep learning framework with the noise inclined module and denoise framework for hyperspectral image classification.
First, we model the spectral signature of hyperspectral image with the physical noise model to describe the high intra-class variance of each class and great overlapping between different classes in the image.
Then, a noise inclined module is developed to capture the physical noise within each object and a denoise framework is then followed to remove such noise from the object.
Finally, the CNN with noise inclined module and the denoise framework is developed to obtain discriminative features and provides good classification performance of hyperspectral image.
Experiments are conducted over two commonly used real-world datasets and the experimental results show the effectiveness of the proposed method. The implementation of the proposed method and other compared methods could be accessed at \url{https://github.com/shendu-sw/noise-physical-framework}.
\end{abstract}


%
\IEEEpeerreviewmaketitle



\section{Introduction}


Hyperspectral images by airborne or spaceborne sensors have played an important role in land cover detection, resource management, quality control, and others \cite{b2, b7}. The hundreds of spectral channels in the image can provide plentiful of spectral information to discriminate different objects of interest in classification task which tries to assign the unique land-cover label to each pixel. In spite of the high spectral resolution in hyperspectral image classification, there exists great noisy in each pixel which would make the great intra-class variance and inter-class similarity. This multiplies the difficulty to extract discriminative features from the image and leads to the degradation of the classification performance. Therefore, exploring an effective and proper model to remove noise from the image is urgent to perform effective image classification.

Nowadays, deep learning based methods have been introduced into hyperspectral image classification and contributed to dramatic improvements \cite{b4}. Deep learning networks (DNNs), especially convolutional neural networks (CNNs), can capture both the local and global information and thus provide discriminative features.
To further improve the potential representational ability of CNNs to extract spatial and spectral features for hyperspectral image classification, most of prior works commonly focus on deepen or widen their network architectures \cite{b14} through the utilization of multi-scale information, spatial information, rotation variance, and others.
For instance,
\cite{b10} utilizes the multi-scale information and proposes a multiscale covariance maps for improvement of hyperspectral image classification. Similarly, \cite{b4} also focuses on embedding the multi-scale information into the deep model.
\cite{b11} and \cite{b13} tries to leverage the spatial and spectral information of the hyperspectral image.
\cite{b12} proposes a pyramidal bottleneck residual blocks which gradually increases the convolutional filters of the network to improve the spectral-spatial features extracted by the deep model.
These prior methods pursue good performance over hyperspectral image classification by improving the model on spectral and spatial features while ignore the intrinsic physical characteristics of the image itself.

To utilize these intrinsic physical characteristics, some prior works utilize such physical prior to construct the training loss and supervise the training process of the deep model. As a representative work, \cite{b3} leverages the statistical properties of each class of hyperspectral image and construct the statistical loss for better representation.
Similarly, \cite{b1} directly utilizes the high dimension property and models the hyperspectral image as nonlinear data manifold, and then constructs the training loss under such manifold assumption.
These methods consider the class distributions of the image and construct the corresponding training loss which can supervise the training process and embed such class information with back propagation of the network. Thus good performance can be obtained with these additional information. However, these methods mainly use the class distribution from the view of training loss while ignore the potential of network module to extract these intrinsic physical characteristics.

As for hyperspectral images, each pixel is generated by radiance signals combined with the sensor noise as the adjacency radiance.
These noises bring statistical variance to the spectral signature of pixels from the same class and therefore different classes in the image present great overlapping of the spectral curves.
Such noise characteristics would negatively affect the deep model to discriminate different classes.
To solve the problem, motivated by \cite{b5}, this work proposes a CNN to remove these noise and provide clean features from the image for better classification performance.
Our method includes a noise inclined module to represent the generated noise from the image and a denoise framework to remove such noise. Such framework can provide clean features from the hyperspectral image and significantly improve the performance of the deep model.

To be concluded, the contributions of this work are twofold. First, this work develops a noise inclined module to extract noise information from the hyperspectral image.
The module utilizes the noise space to represent the noise from the image. The diversity regularization and the large scale of noise space can guarantee a good representation of the noise from the image.
Then, this work proposes a CNN framework to remove the extracted noise by the noise inclined module from the features, and provide clean and discriminative features to improve the classification performance.

The remainder of this paper is arranged as follows.
Section \ref{sec:method} details the construction of the proposed framework for hyperspectral image classification.
Experimental results and analysis  over two real-world datasets are summarized in Section \ref{sec:experiments}.
Section \ref{sec:conclusions} concludes this paper with some future directions.

\section{Proposed Method}\label{sec:method}

Denote $X = \{{\bf x}_1, {\bf x}_2, \cdots,{\bf x}_N\}$ as the set of training samples of the hyperspectral image where $y_i$ is the corresponding label of the sample ${\bf x}_i$ and $N$ is the number of training samples.

\begin{figure*}[htp]

  \centering
 {\includegraphics[width=0.98\textwidth]{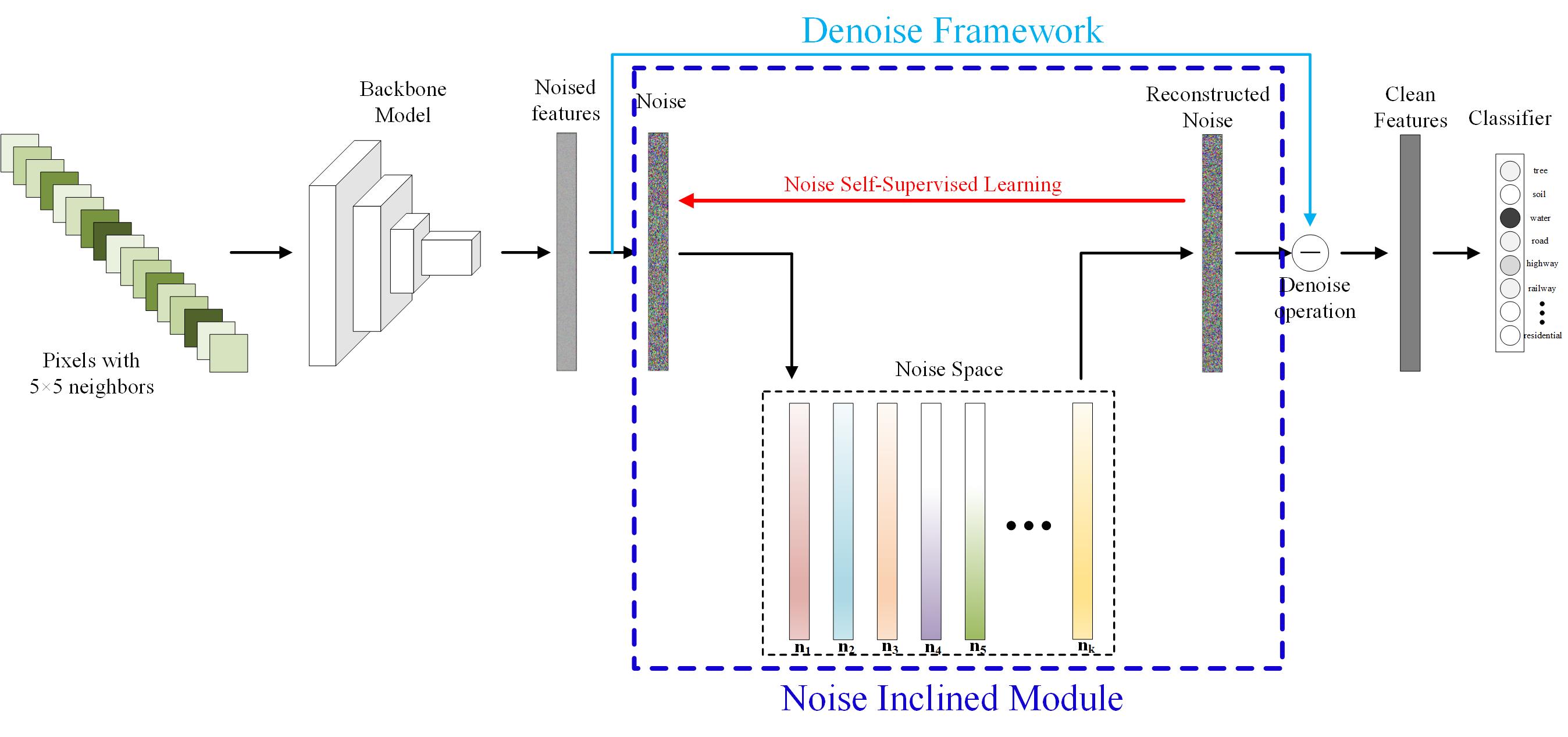}}
\hfill
\caption{Flowchart of the proposed method for hyperspectral image classification. }
\label{fig:flowchart}
\end{figure*}

\subsection{Physical Modelling}

The hyperspectral image can be physically modelled by at-sensor radiance signal model. Generally, the total radiation signal reaching the sensor consists of three components, namely the reflected radiance (radiation reflected from the pixel of interest), the  adjacency radiance (radiation reflected from the surface surrounding the pixel of interest and scattered by the air volume into the sensor) and other radiance from sensor noise.
The reflected radiance generates the spectral signature of a certain object and the other radiances formulate the noise imposed on the object.

Therefore, for hyperspectral image processing, the spectral characteristics of each object
can be defined by a single class spectrum or a target subspace combined with noise from the neighbors, sensors, and others \cite{b8, b9}. Let ${\bf x}_\Lambda^{m}$ denote the object of interest from $m$-th class and it can be formulated as
\begin{equation}\label{eq:01}
{\bf x}_\Lambda^{m}={\bf s}^{m}+{\bf n}({\bf x}_\Lambda^{m})
\end{equation}
where ${\bf s}^{m}$ is the spectral signature of the $m$-th class, and ${\bf n}({\bf x}_\Lambda^{m})$ is the random additive  noise from the neighbors and sensors imposed on this specific object.

To represent these noises imposed on the image,
this work denotes the noise space $S=\{{\bf n}_1, {\bf n}_2, \cdots, {\bf n}_k\}$ where ${\bf n}_i$ stands for $i^{th}$ base noise of the space.
Then, the noise can be seen as the linear weighted of noises from multiple sources in noise space and it can be calculated as
\begin{equation}\label{eq:02}
{\bf n}({\bf x}_\Lambda^{m})=\lambda_1({\bf x}_\Lambda^{m}){\bf n}_1+\lambda_2({\bf x}_\Lambda^{m}){\bf n}_2+\cdots+\lambda_k({\bf x}_\Lambda^{m}){\bf n}_k
\end{equation}
where ${\bf n}_1, {\bf n}_2,\cdots, {\bf n}_k$ represents the base noise from different sources and $\lambda_i({\bf x}_\Lambda^{m})$ denotes the weight parameter of the noise ${\bf n}_i(i=1,2,\cdots, k)$.

\subsection{Noise Inclined Module}

This subsection proposes the noise inclined module to learn these base noises adaptively, estimate the weight parameters, and provide the representation of the noise of each object in the noise space.
The noise inclined module consists of three parts, namely the noise extraction module to extract noise from the features, the noise reconstruction mudule to reconstruct the noise in the noise space and the self-supervised module to update the noise space.

\subsubsection{Noise Extraction Module}
The noise extraction module is performed by the noise mapping operation.
In general imaging process, the noise ${\bf n}({\bf x}_\Lambda^{m})$ of ${\bf x}_\Lambda^{m}$ is highly related to the signature ${\bf s}^{m}$, which can be represented as
${\bf n}({\bf x}_\Lambda^{m})\propto {\bf s}^{m}$.
Based on Eq. \ref{eq:01}, the noise is also related to the noised feature ${\bf x}_\Lambda^{m}$, 
\begin{equation}
{\bf n}({\bf x}_\Lambda^{m})\propto {\bf x}_\Lambda^{m}
\end{equation}
Considering the imaging mechanism, ${\bf n}({\bf x}_\Lambda^{m})$ can be calculated through the non-linear transformation of ${\bf x}_\Lambda^{m}$, which is implemented by the fully-connected layer in this work.
Denote $\phi({\bf x}_\Lambda^{m})$ as the learned features from the CNN where $\phi(\cdot)$ represents the mapping function of the CNN. Then, the extracted noise can be calculated by
\begin{equation}
{\bf n}_f({\bf x}_\Lambda^{m})={\bf W}\phi({\bf x}_\Lambda^{m})+b
\end{equation}
where ${\bf W}, b$ represent the weight and the bias of the layer, respectively.

\subsubsection{Noise Reconstruction Module}
The key process for noise reconstruction is to reconstruct the noise in the noise space based on the extracted noise from noise extraction module.  
The noise reconstruction process can be seen as the following optimization,
\begin{equation}\label{eq:optimization}
\min\limits_{\lambda_1,\cdots, \lambda_k}L_u=L_r+L_s+\alpha L_d
\end{equation}
where $L_r=\|{\bf n}_f({\bf x}_\Lambda^{m})-\sum_{i=1}^k\lambda_i{\bf n}_i\|$ denotes the reconstruction term, $L_s=\sum_{i=1}^k|\lambda_i({\bf x}_\Lambda^{m})|$ is the regularization term to enforce the unique reconstruction solution, and $L_d$ is the noise diversity promoting augment and $\alpha$ is a constant that controls tradeoff between task objective and diversity.
This paper defines a angle-based difference measure for diversity, that is,
\begin{equation}
L_d=\frac{1}{k(k-1)}\sum\limits_{l\neq m}{\bf n}_l^T{\bf n}_m
\end{equation}
It should be noted that $L_d$ mainly performs for the update of ${\bf n}_i$.

Instead of obtaining the values of $\lambda_i({\bf x}_\Lambda^{m})(i=1,\cdots,k)$ through optimization methods, this work tries to estimate these values directly using the correlation between the base noise and the noise features. Generally, the base noises ${{\bf n}_1, \cdots, {\bf n}_k}$ are expected to be indepedent to each other, therefore, this work tries to use the cosine similarity to estimate the values and impose the indepedent regularization over the base noises. 

As \cite{b5}, we compute the cosine similarity $s_{j}({\bf x}_\Lambda^{m})$ of the $j$-th base noise ${\bf n}_j$ and the extracted noise $n_f({\bf x}_\Lambda^{m})$.
\begin{equation}
s_{j}({\bf x}_\Lambda^{m})=\frac{{\bf n}_j{\bf n}_f({\bf x}_\Lambda^{m})}{\|{\bf n}_j\|\|{\bf n}_f({\bf x}_\Lambda^{m})\|}
\end{equation}
Then, the pre-reconstructed noise is calculated by
\begin{equation}
{\bf n}'=\sum_{j=1}^ks_{j}({\bf x}_\Lambda^{m}){\bf n}_i
\end{equation}

To preserve the energy of the noise, the final estimation of $\lambda_j$ given ${\bf x}_\Lambda^{m}$ can be measured by 
\begin{equation}
\lambda_j({\bf x}_\Lambda^{m})=\frac{\|{\bf n}_f({\bf x}_\Lambda^{m})\|}{\|{\bf n}'\|}s_{j}({\bf x}_\Lambda^{m})
\end{equation}

Then, the final reconstructed noise can be formulated as
\begin{equation}
{\bf n}_{res}({\bf x}_\Lambda^{m})=\sum_{j=1}^k\lambda_j({\bf x}_\Lambda^{m}){\bf n}_j
\end{equation}

\subsubsection{Noise Self-Supervised Module}

These base noises in noise space are updated through self-supervised learning
based on the gradient descent of Eq. \ref{eq:optimization} and it can be calculated as
\begin{equation}
\frac{\partial L_u}{\partial {\bf n}_i}=\frac{\partial L_s}{\partial {\bf n}_i}+\alpha\frac{\partial L_d}{\partial {\bf n}_i}
\end{equation}
Based on derivation rule,
\begin{equation}
\frac{\partial L_s}{\partial {\bf n}_i}=-2\lambda_i({\bf x}_\Lambda^{m})({\bf n}_f({\bf x}_\Lambda^{m})-\sum_{j=1}^k\lambda_j({\bf x}_\Lambda^{m}){\bf n}_j)
\end{equation}
Besides, $\displaystyle{\frac{\partial L_d}{\partial {\bf n}_i}}$ is used to decrease the redundancy and make independent between different base noises, and it can be calculated by
\begin{equation}
\frac{\partial L_d}{\partial {\bf n}_i}=\frac{2}{k(k-1)}\sum\limits_{l\neq i}{\bf n}_l
\end{equation}

In addition, this work uses the decay way for self-supervised update.
The final self-supervised update of the base noise can be formulated as
\begin{equation}
{\bf n}_j^{(k+1)}=\beta {\bf n}_j^{(k)} + (1-\beta) \times \frac{\partial L_u}{\partial {\bf n}_i}
\end{equation}
where  $\beta\in [0,1]$ is the decay rate. This strategy is also the self-supervised updating just as \cite{b5}.

\subsection{The Proposed Framework}

This work proposes a novel framework with noise inclined module and denoise operation to extract clean and discriminative features from the hyperspectral image.
The noise inclined module is used to extract the physical noise from the image, reconstruct the noise in a noise space, and finally update the noise space under self-supervised way.

Then, the denoise operation is followed to provide the clean features from the image, and the denoise operation can be implemented by
\begin{equation}
f_{clean}({\bf x}_\Lambda^{m})=\phi({\bf x}_\Lambda^{m})-{\bf n}_{res}({\bf x}_\Lambda^{m})
\end{equation}
The obtained clean features are then used for the classification of different objects in the image.
The flowchart of the proposed method is shown in Fig. \ref{fig:flowchart}.

This work utilizes the center loss \cite{b15} for the training of the deep model to increase the inter-class variance of the learned clean features, and the loss can be formulated as
\begin{equation}
L_C=\frac{1}{2}\sum\limits_{{\bf x}_\Lambda^{m}\in X}\|f_{clean}({\bf x}_\Lambda^{m})-{\bf c}_{m}\|^2_2
\end{equation}
where ${\bf c}_{m}$ denotes the class center in the loss.

\section{Experimental Results}\label{sec:experiments}

\subsection{Experimental Setups}

To validate the effectiveness of the proposed method, we conduct experiments over two real-world  hyperspectral image datasets, namely Pavia University dataset and Houston 2013 dataset.

{\bf Pavia University data \cite{b18}} was collected by the reflective optics system imaging spectrometer (ROSIS-3) sensor ranging from 0.43 to 0.86 $\mu m$ with a spatial resolution of 1.3$m$ per pixel. The data contains $610\times 340$ pixels with 115 bands where 12 bands have removed due to noise and absorption and the remaining 103 channels are used. 42776 samples from 9 classes have been labeled for experiments. 

{\bf Houston 2013 data \cite{b16}} was acquired over the University of Houston campus and the neighboring urban area with a resolution of 2.5 m/pixel, and provided as part of the 2013 IEEE Geoscience and remote Sensing Society data fusion contest.
it consists of $349\times 1905$ pixels of which a total of 15011 labeled samples divided into 15 classes have been chosen for experiments. Each pixel denotes a sample and contains 144 spectral bands ranging from 0.38 to 1.05 $\mu m$. 

Pytorch is used as the deep learning framework and all the experiments are carried out on a PC with an Intel@ Xeon(R) Gold 6226R CPU and Nvidia Quadro RTX 6000 GPU. 
For pavia university data, 200 samples per class are randomly selected for training and the remainder are for testing while for Houston2013 data, the training and the testing set are following the 2013 GRSS datafusion contest.
The dimension of the learned features is set to 400.
The learning rate is set to $1e^{-4}$ and the training batch is set to 4. The hyperparameter $\alpha$, $\beta$ is set to $1$, 0.9, respectively.

\begin{figure*}[t]
\centering
 \includegraphics[width=0.99\linewidth]{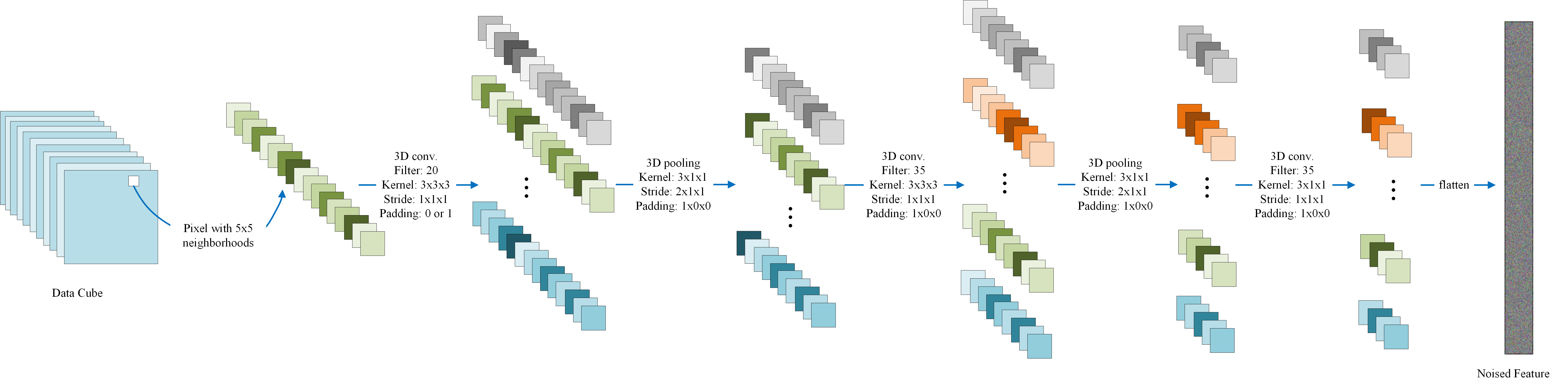}
   \caption{CNN model adopted in the experiments to implement the proposed method for hyperspectral image classification.}
\label{fig:houston2013}
\end{figure*}

%

\subsection{General Performance}

At first, we present the general performance of the proposed method for hyperspectral image classification. In this set of experiments, the number of base noise in the noise space is set to 1024. The backbone of the proposed method is the 3-D CNN.

As showed in table \ref{table:pavia} and \ref{table:houston2013}, the proposed method can improve the performance of vanilla CNNs by introducing the physical information in the training process. As for the Pavia University data and the Houston2013 data, the CNNs with the proposed framework can significantly improve the CNNs' performance. For Pavia University, the accuracy can increase from $94.97\%\pm 0.69\%$ to $96.56\%\pm 0.47\%$ with the proposed framework. For Houston2013 data, the accuracy can be improved by about $6\%$. This indicates that the  physical information is effective for the training of the deep CNNs in the hyperspectral image classification literature.

\begin{figure}[t]
\centering
\subfigure[]{\label{subfig:pavia_cnn}\includegraphics[width=0.32\linewidth]{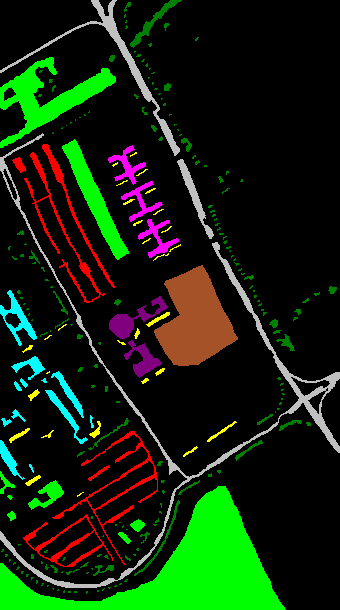}}
 \subfigure[]{\label{subfig:pavia_cnn}\includegraphics[width=0.32\linewidth]{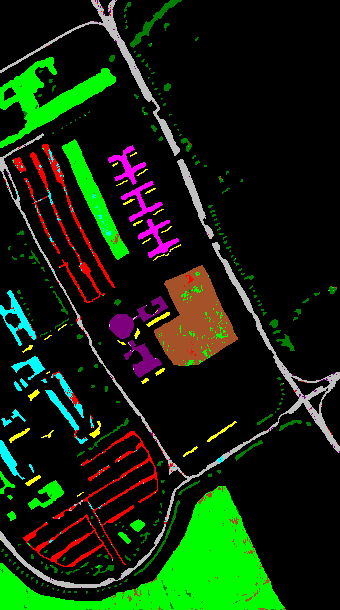}}
\subfigure[]{\label{subfig:pavia_cnn}\includegraphics[width=0.32\linewidth]{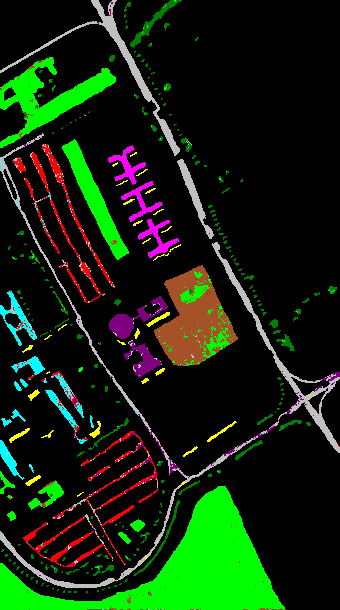}}
\subfigure[]{\label{subfig:pavia_cnn}\includegraphics[width=0.32\linewidth]{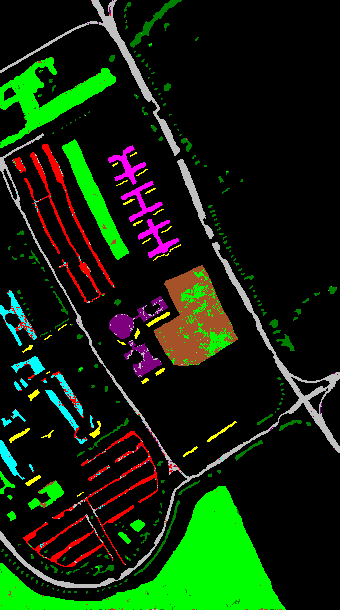}}
\subfigure[]{\label{subfig:pavia_cnn}\includegraphics[width=0.32\linewidth]{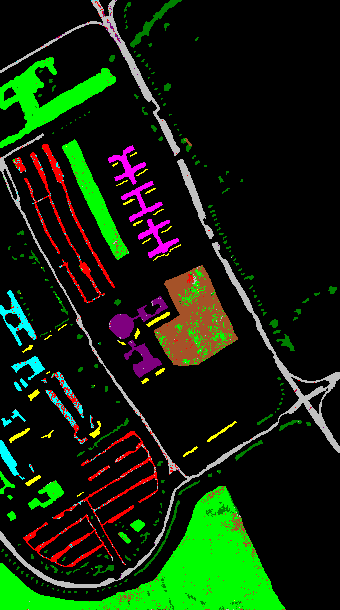}}
 \subfigure[]{\label{subfig:pavia_proposed}\includegraphics[width=0.32\linewidth]{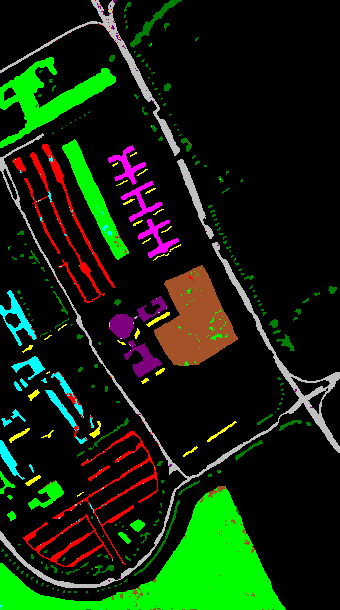}}

   \caption{Classification mapping by different methods over Pavia University (Overall accuracies). (a) Ground truth; (b) 3-D CNN (95.42\%); (c) HybridSN(46.92\%); (d) PResnet(94.16\%); (e) SSFTTnet(92.72\%); (f) Proposed Method(97.24\%).}
\label{fig:pavia}
\end{figure}

Furthermore, Fig. \ref{fig:pavia} and \ref{fig:houston2013} shows the classification maps of different datasets. It also demonstrates that the proposed method using the physical information can remarkably decrease the classification errors  and further improve the classification performance.

\begin{table*}[t]
\begin{center}
\caption{Classification accuracies ($Mean\pm SD$) (OA, AA, and Kappa) of different methods achieved on the Pavia University data.}
\label{table:pavia}
\begin{tabular}{|c | c | c c c c c c|}
\hline
\multicolumn{2}{|c|}{\bf Methods}     &  {\bf SVM-POLY} &  {\bf 3-D CNN}&  {\bf HybridSN}&  {\bf PResNet} &  {\bf SSFTTNet} &  {\bf Proposed Method} \\
\hline\hline
\multirow{9}{*}{\rotatebox{90}{\tabincell{c}{\textbf{Classification} \\ \textbf{Accuracies (\%)}}}}          & C1   &  $82.79\pm 1.42$  &  $95.59\pm 1.15$ &$91.52\pm 1.76$ & $\mathbf{97.10\pm0.87}$& $94.97\pm1.95$&  ${95.92\pm 1.56}$\\
                                                                                                             & C2   &  $85.91\pm 0.74$  &  $96.82\pm 0.69$ & $96.20\pm2.65$& $97.04\pm1.85$& $95.92\pm1.81$&  $\mathbf{97.46\pm 0.74}$\\
                                                                                                             & C3   &  $86.12\pm 0.34$  &  $89.68\pm 2.12$ & $70.56\pm9.31$& $84.20\pm4.69$& $77.36\pm0.37$&  $\mathbf{92.22\pm 1.07}$\\
                                                                                                             & C4   &  $97.75\pm 0.12$  &  $97.60\pm 1.30$ & $92.32\pm3.51$& $97.91\pm0.74$& $96.23\pm2.77$&  $\mathbf{98.03\pm 0.92}$\\
                                                                                                             & C5   &  $99.65\pm 0.12$  &  $100.0\pm 0.00$ & $99.91\pm0.12$& $100.0\pm0.00$& $100.0\pm0.00$&  $\mathbf{100.0\pm 0.00}$\\
                                                                                                             & C6   &  $91.81\pm 0.25$  &  $86.77\pm 4.42$ & $85.71\pm3.57$& $85.39\pm8.62$& $74.61\pm11.74$&  $\mathbf{94.70\pm 3.28}$\\
                                                                                                             & C7   &  $94.56\pm 0.06$  &  $97.36\pm 0.92$ & $84.16\pm12.89$& $90.18\pm5.76$& $95.18\pm3.69$&  $\mathbf{97.72\pm 1.08}$\\
                                                                                                             & C8   &  $87.84\pm 1.08$  &  $92.64\pm 2.19$ & $84.92\pm10.60$& $89.82\pm0.75$& $81.40\pm9.32$&  $\mathbf{94.47\pm 2.05}$\\
                                                                                                             & C9   &  ${\mathbf{100.0\pm 0.00}}$  &  $99.71\pm 0.31$ & $99.46\pm0.76$& $99.93\pm0.09$& $99.40\pm0.09$&  ${99.84\pm 0.22}$\\
 \hline
 \multicolumn{2}{|c|}{{\bf OA}  (\%)}      &  $88.00\pm 0.19$ &  $94.97\pm 0.69$  & $91.64\pm1.16$& $94.47\pm0.44$& $91.34\pm1.95$&  $\mathbf{96.56\pm 0.47}$\\
 \hline
 \multicolumn{2}{|c|}{{\bf AA}  (\%)}      &  $91.83\pm 0.06$ &  $95.13\pm 0.60$  & $89.42\pm1.67$& $93.51\pm1.83$& $90.56\pm3.04$&  $\mathbf{96.70\pm 0.55}$\\
 \hline
 \multicolumn{2}{|c|}{{\bf KAPPA} (\%)}    &  $84.29\pm 0.23$ &  $93.32\pm 0.89$  & $88.96\pm1.46$& $92.66\pm0.61$& $88.58\pm2.58$&  $\mathbf{95.41\pm 0.63}$\\
\hline
\end{tabular}
\end{center}
\end{table*}

\begin{table*}[t]
\begin{center}
\caption{Classification accuracies ($Mean\pm SD$) (OA, AA, and Kappa) of different methods achieved on the Houston 2013 data.}
\label{table:houston2013}
\begin{tabular}{|c | c | c c c c c c|}
\hline
\multicolumn{2}{|c|}{\bf Methods}     &  {\bf SVM-POLY} &  {\bf 3-D CNN}&  {\bf HybridSN}&  {\bf PResNet} &  {\bf SSFTTNet} &  {\bf Proposed Method} \\
\hline\hline
\multirow{15}{*}{\rotatebox{90}{\tabincell{c}{\textbf{Classification} \\ \textbf{Accuracies (\%)}}}}          & C1   &  $81.48 $  &  $\mathbf{83.84\pm 2.56}$ &$80.10\pm2.75$ & $82.24\pm1.07$&$82.29\pm0.34$ &  ${82.85\pm 0.08}$\\
                                                                                                             & C2   &  $81.86 $  &  $83.57\pm 2.54$ & $82.94\pm0.33$& $83.88\pm1.53$& $89.24\pm8.31$&  $\mathbf{90.04\pm 4.12}$\\
                                                                                                             & C3   &  ${99.41} $  &  $95.21\pm 2.13$ & $91.49\pm0.84$&$97.03\pm0.00$ &$\mathbf{99.70\pm0.14}$ &  ${96.48\pm 0.84}$\\
                                                                                                             & C4   &  $92.61 $  &  $94.34\pm 2.59$ & $91.10\pm2.54$&$93.04\pm0.47$ & $97.35\pm0.67$&  $\mathbf{97.99\pm 1.32}$\\
                                                                                                             & C5   &  $96.88 $  &  $98.48\pm 0.88$ & $98.72\pm1.54$& $98.72\pm0.33$&$\mathbf{99.62\pm0.00}$ &  ${98.92\pm 0.46}$\\
                                                                                                             & C6   &  $93.71 $  &  $91.19\pm 2.45$ & $82.17\pm4.45$& $96.15\pm1.48$& $\mathbf{97.90\pm2.97}$&  ${94.83\pm 2.82}$\\
                                                                                                             & C7   &  $80.22 $  &  $83.71\pm 5.16$ & $78.31\pm0.73$&$85.49\pm4.02$ & $73.23\pm8.31$&  $\mathbf{90.54\pm 1.48}$\\
                                                                                                             & C8   &  $36.28 $  &  $63.84\pm 2.21$ & $67.47\pm3.42$&$66.81\pm0.07$ &$\mathbf{81.48\pm17.86}$ &  ${70.65\pm 3.21}$\\
                                                                                                             & C9   &  $72.90 $  &  $69.63\pm 4.58$ & $68.13\pm0.87$&$\mathbf{76.11\pm0.53}$ &$69.88\pm17.76$ &  ${73.82\pm 3.87}$\\
                                                                                                             & C10   &  $55.89 $  &  $49.44\pm 4.04$ & $56.56\pm1.64$& $51.59\pm4.03$&$\mathbf{60.38\pm30.78}$ &  ${59.13\pm 3.01}$\\
                                                                                                             & C11   &  $\mathbf{75.14} $  &  $62.52\pm 3.35$ &$72.01\pm1.88$ &$73.29\pm1.41$ &$67.47\pm32.70$ &  ${74.91\pm 2.60}$\\
                                                                                                             & C12   &  $72.24 $  &  $69.34\pm 7.59$ &$77.86\pm9.17$ &$71.37\pm4.08$ &$62.58\pm1.83$ &  $\mathbf{88.49\pm 1.59}$\\
                                                                                                             & C13   &  $72.28 $  &  $91.58\pm 2.90$ &$88.25\pm3.72$ &$90.70\pm2.23$ &$91.58\pm0.99$ &  $\mathbf{91.16\pm 1.52}$\\
                                                                                                             & C14   &  $\mathbf{100.0} $  &  $96.68\pm 0.78$ &$99.19\pm1.15$ &$98.99\pm0.29$ &$100.0\pm0.00$ &  ${94.98\pm 1.02}$\\
                                                                                                             & C15   &  $\mathbf{98.31} $  &  $93.66\pm 1.97$ &$96.83\pm3.29$ &$93.45\pm6.88$ &$96.41\pm0.90$ &  ${94.88\pm 2.87}$\\
 \hline
 \multicolumn{2}{|c|}{{\bf OA}  (\%)}      &  $77.24 $ &  $78.40\pm 1.37$  &$79.46\pm0.21$ &$80.61\pm0.72$ &$80.96\pm1.44$ &  $\mathbf{84.41\pm 0.73}$\\
 \hline
 \multicolumn{2}{|c|}{{\bf AA}  (\%)}      &  $80.61 $ &  $81.80\pm 1.03$  &$82.08\pm0.12$ & $83.93\pm1.00$&$84.61\pm1.24$ &  $\mathbf{86.64\pm 0.76}$\\
 \hline
 \multicolumn{2}{|c|}{{\bf KAPPA} (\%)}    &  $75.51 $ &  $77.00\pm 1.43$  &$78.10\pm0.21$ &$79.32\pm0.76$ &$79.69\pm1.51$ &  $\mathbf{83.31\pm 0.76}$\\
\hline
\end{tabular}
\end{center}
\end{table*}

\begin{figure}[t]
\centering
 
 \subfigure[]{\label{subfig:houston2013_tslabel}\includegraphics[width=0.9\linewidth]{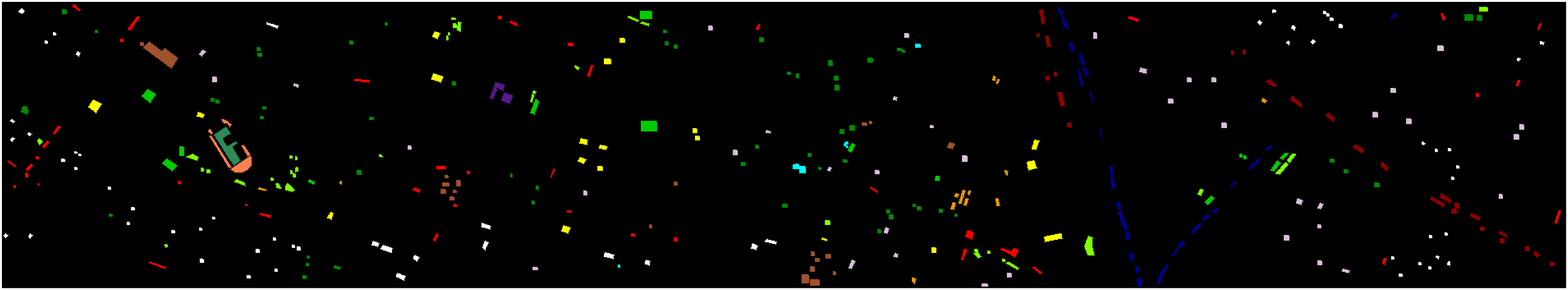}}
\subfigure[]{\label{subfig:pavia_neighbor}\includegraphics[width=0.9\linewidth]{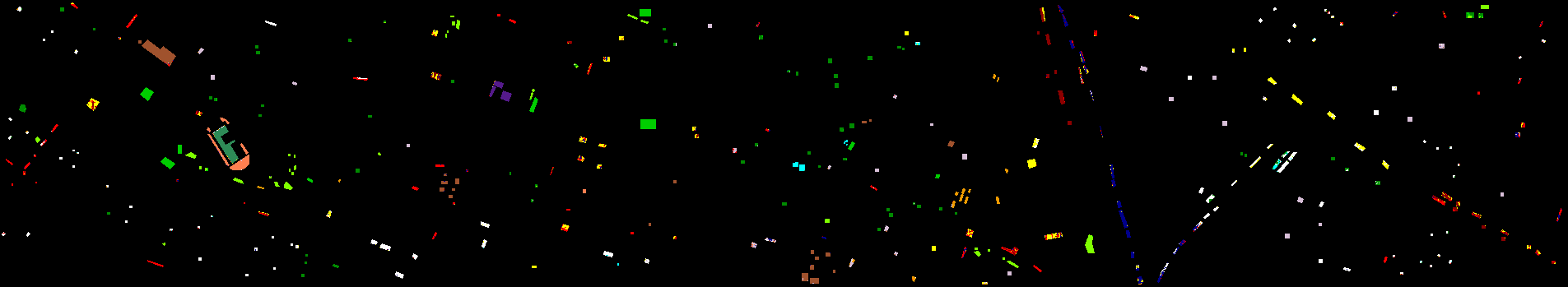}}
\subfigure[]{\label{subfig:pavia_neighbor}\includegraphics[width=0.9\linewidth]{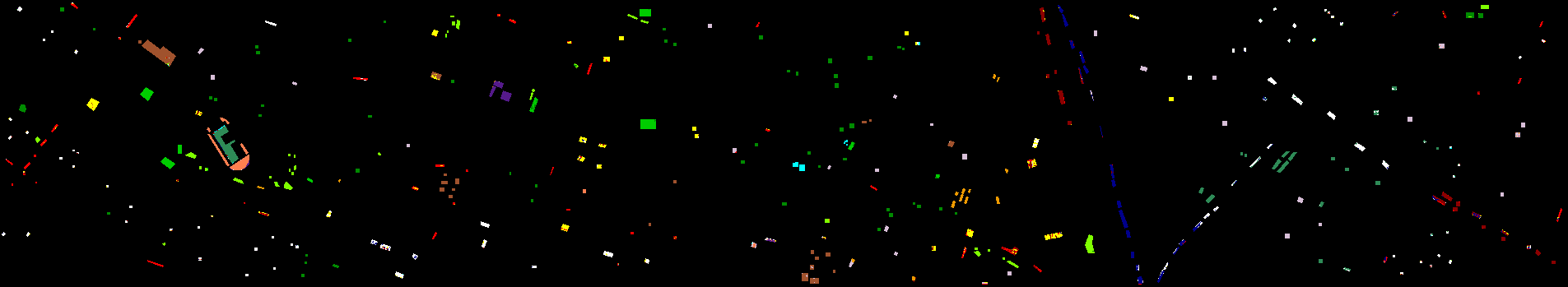}}
\subfigure[]{\label{subfig:pavia_neighbor}\includegraphics[width=0.9\linewidth]{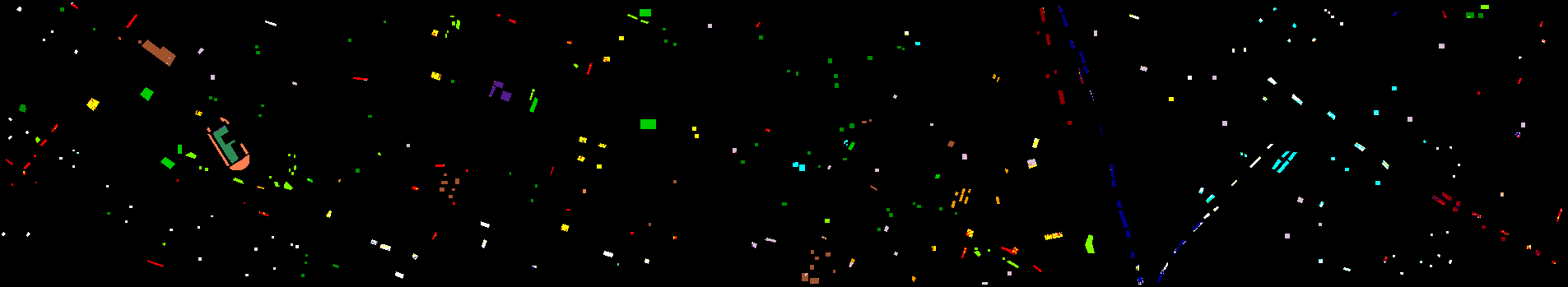}}
\subfigure[]{\label{subfig:pavia_neighbor}\includegraphics[width=0.9\linewidth]{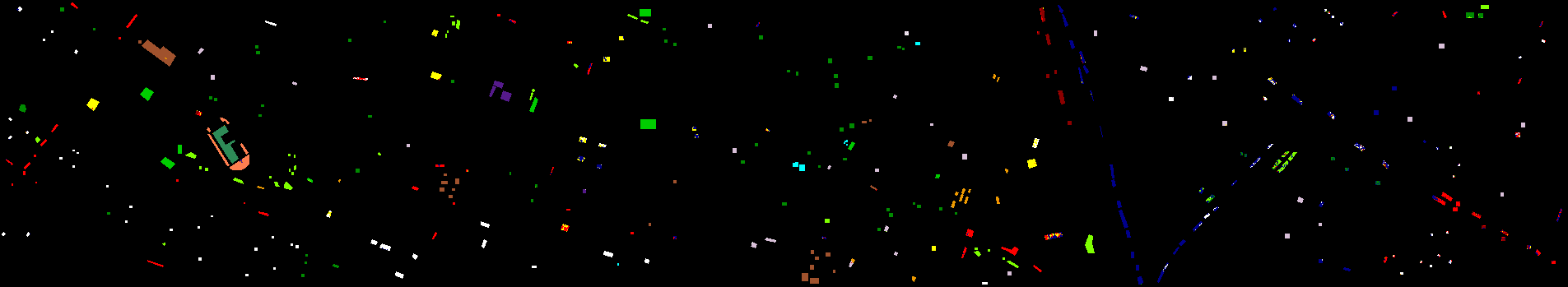}}
\subfigure[]{\label{subfig:pavia_neighbor}\includegraphics[width=0.9\linewidth]{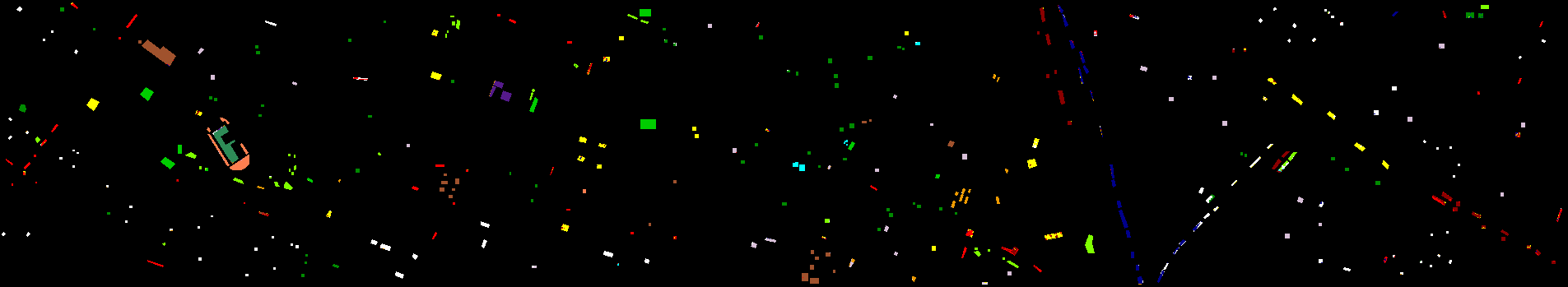}}

   \caption{Classification mapping by different methods over Houston2013 (Overall accuracies). (a) Ground truth; (b) 3-D CNN (77.28\%); (c) HybridSN(79.31\%); (d) PResnet(81.12\%); (e) SSFTTnet(81.98\%); (f) Proposed Method(85.46\%).}
\label{fig:houston2013}
\end{figure}


\subsection{Performance with Different Number of Base Noise}

\begin{figure}[t]
\centering
 \subfigure[]{\label{subfig:pavia_number}\includegraphics[width=0.48\linewidth]{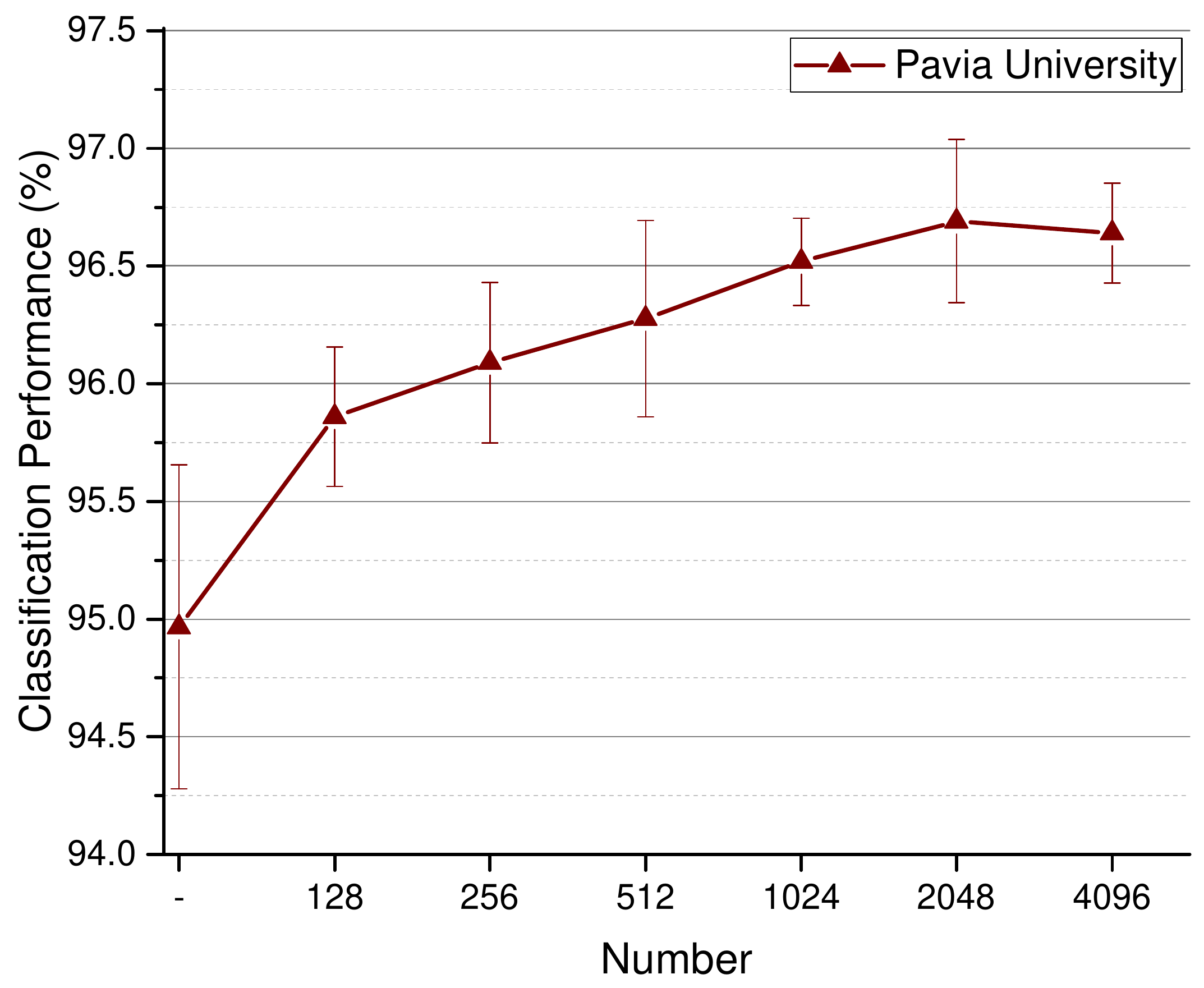}}
 \subfigure[]{\label{subfig:houston2013_number}\includegraphics[width=0.48\linewidth]{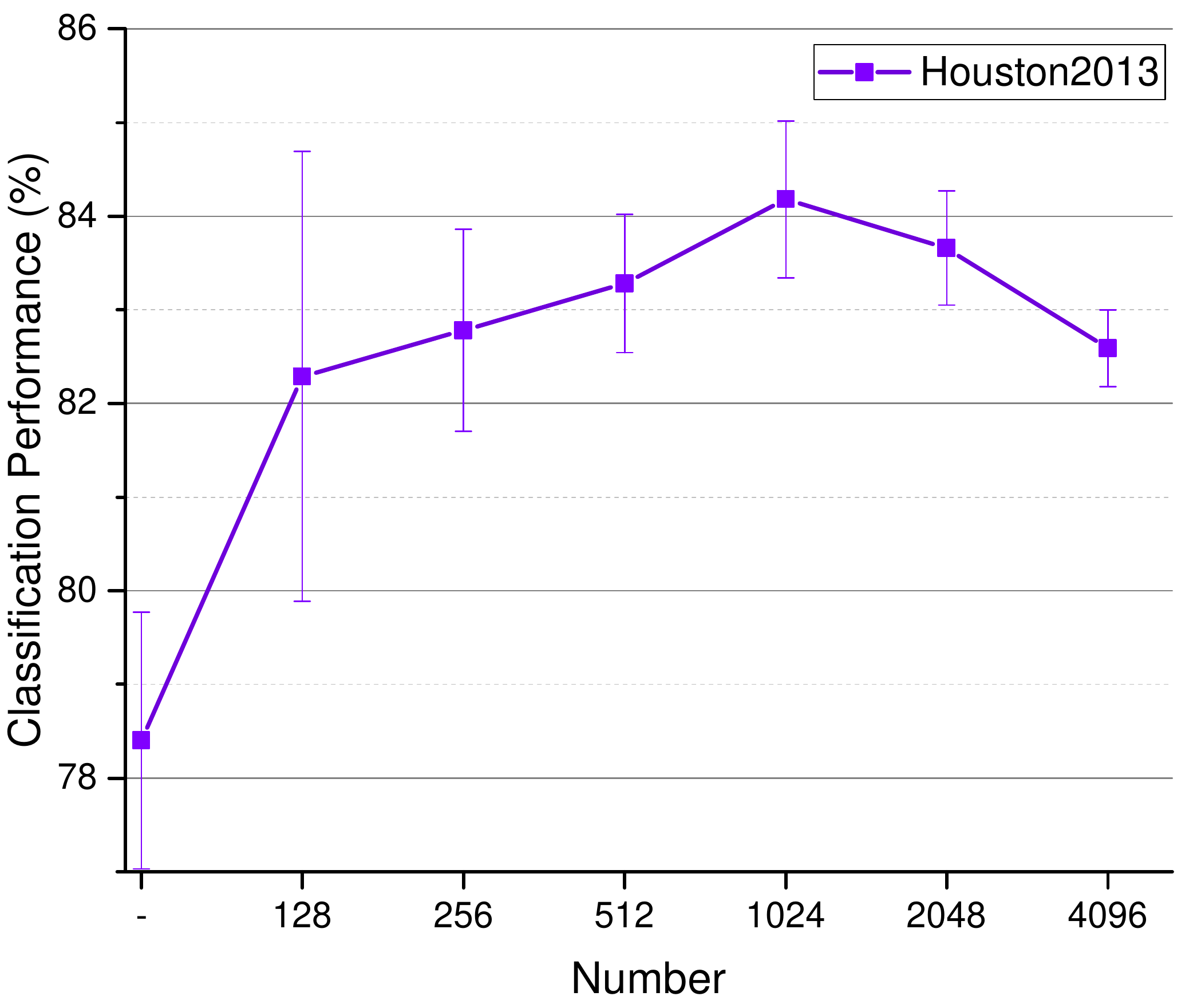}}

   \caption{Classification performance with different number of base noise in the noise space over  (a) Pavia University; (b) Houston2013.}
\label{fig:number}
\end{figure}

In this set of experiments, we test the performation of the proposed framework under different number of base noise in the noise space. 
The number of base noise in the noise space denotes the representational ability of the noise space. Generally, a larger number of base noise can provide a better representation of the noise in the signature characteristics of the hyperspectral image and thus the clean feature can be more discriminative. However, excessive number of base noise increases the learned parameters and makes redundancy between the base noise which makes the degradation of the classification performance. Figure \ref{fig:number} shows the tendencies of the performance under different number of base noise. As for Pavia University data, the classification performance achieves the best with base noise of 2048 while for Houston2013 data, the classification performance achieves best under 1024 base noise. The tendencies of the classification performance over the two datasets validate the conclusions before.

\subsection{Performance with Different Sizes of neighbors}

\begin{figure}[t]
\centering
 \subfigure[]{\label{subfig:pavia_neighbor}\includegraphics[width=0.48\linewidth]{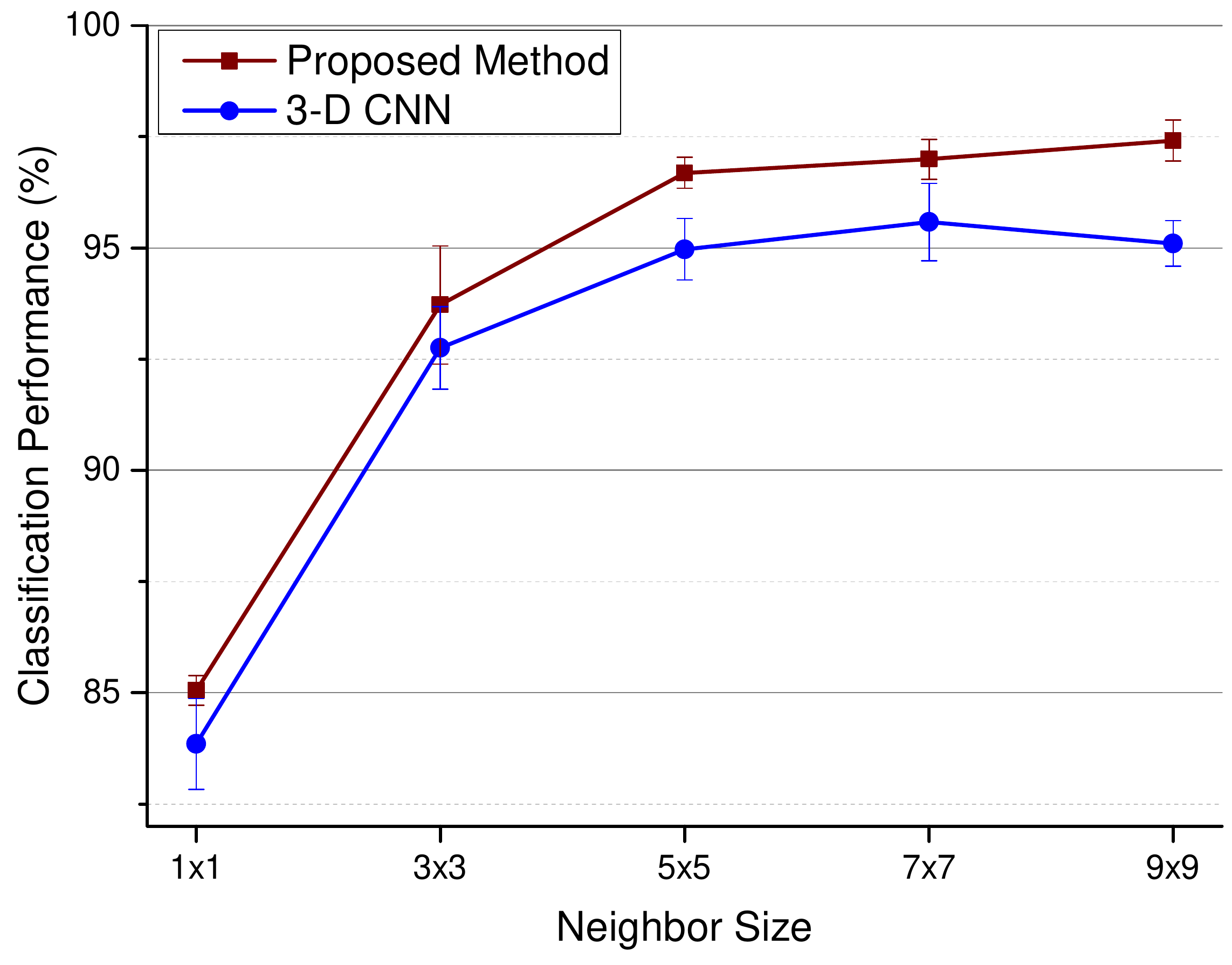}}
 \subfigure[]{\label{subfig:houston2013_neighbor}\includegraphics[width=0.48\linewidth]{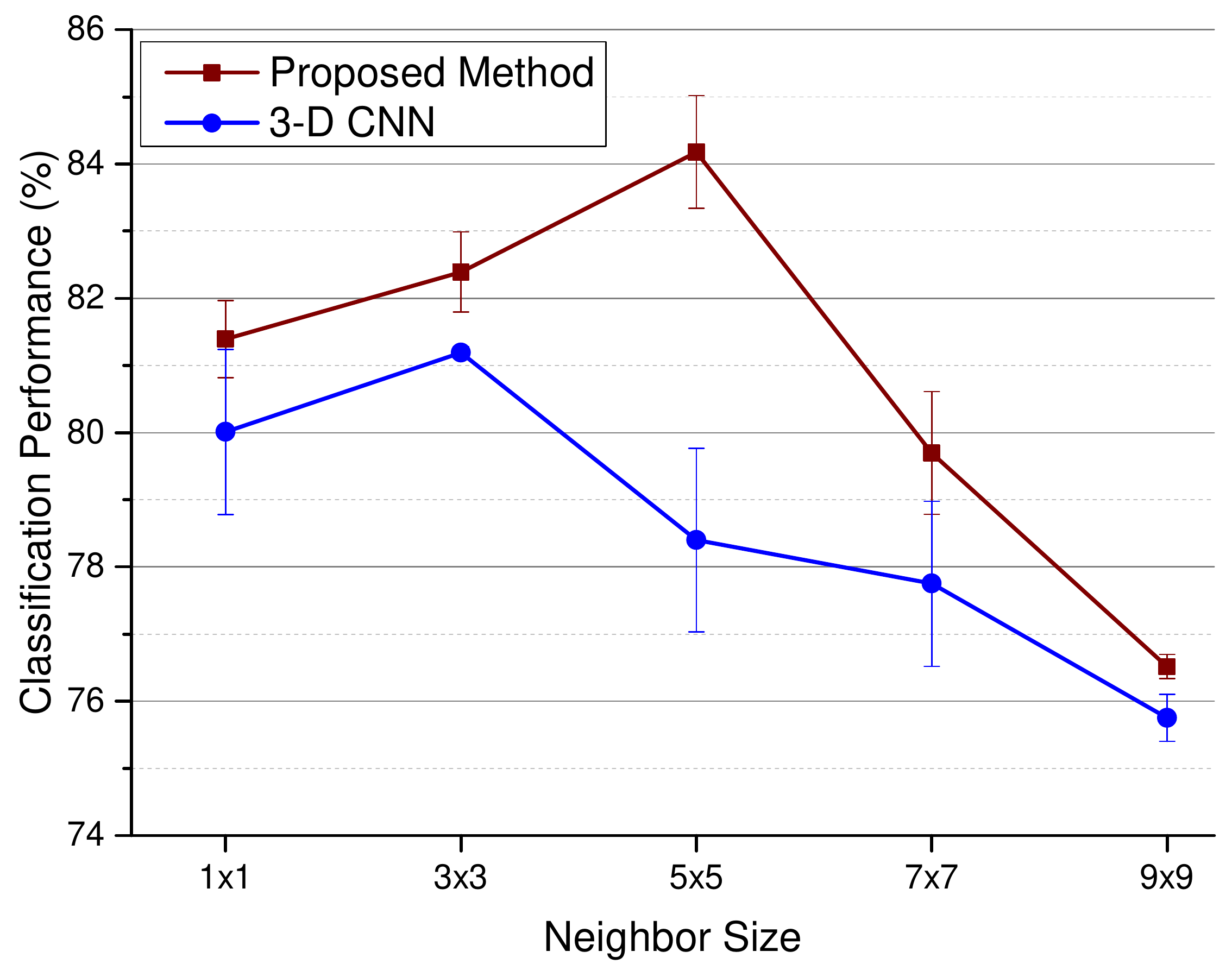}}

   \caption{Classification performance with different neighbor sizes over  (a) Pavia University; (b) Houston2013.}
\label{fig:neighbor}
\end{figure}

In this set of experiments, we presents the performance under different sizes of neighbors. Fig \ref{fig:neighbor} shows the tendencies with different neighbor sizes over the two datasets. As the figure shows,
the 3-D CNN is not very friendly to extract features from hyperspectral image with larger neighbor sizes. Generally, spectral-spatial classification with larger neighbor size contains more spatial information and thus provide more discriminative features for classification. While due to the limitation of the backbone model, the classification performance even decreases with excessive large spatial sizes (e.g. $9\times 9$).

More importantly, as Fig. \ref{fig:neighbor} shows, the classification performance can be significantly improved with the proposed method by using physical information under different sizes of neighbors. Especially, for Pavia University data, the classification accuracy can be increased by 1.72\% under $5\times 5$ neighbor and 2.32\% under $9\times 9$ neighbor. While for Houston2013 data, the classification accuracy can even be increased by about 6\% under $5\times 5$ neighbor.

\subsection{Comparisons with Other Methods}

To thuroughly evaluate the effectiveness of the proposed method, this work selects Support Vector Machine(SVM), 3-D CNN \cite{b2}, PResNet \cite{b19}, HybridSN \cite{b20}, SSFTTNet \cite{b21} as baselines. Table \ref{table:pavia} and \ref{table:houston2013} show the comparisons over the two data sets, respectively. The experimental results in each table are with the same experimental setups and $5\times 5$ neighbors are used as the spatial information in the experiments. 

From table \ref{table:pavia}, we can find that the proposed method which achieves $96.56\%\pm 0.47\%$ OA outperforms these baseline methods over Pavia University data where the best provides an OA of $94.97\%\pm 0.69\%$ by 3-D CNN. As for Houston2013 data, it can be noted that the proposed method also provides an superior performance ($84.41\%\pm 0.73\%$ by proposed method vs $80.96\%\pm 1.44\%$ by SSFTTNet). Furthermore, the proposed method can provide a better performance by using a different backbone CNN. This indicates that the proposed method can provide an comparable or better performance than other recent methods.

\section{Conclusions}\label{sec:conclusions}

In this work, a novel CNN with noise inclined module and denoise framework is developed for hyperspectral image classification.
First, the physical noise model is constructed to represent the noise information. Then, the noise inclined module is developed to adaptively learn the noise from each object. Finally, the denoise framework is followed to remove the extracted noise information and privide clean features to better discriminate different objects. The experimental results over two challenging datasets have demonstrated the effectiveness of the proposed method.

In future work, it is suggested investigating other effective architecture using the intrinsic information.
Another important future avenue of exploration is other intrinsic structures of the hyperspectral image for effective learning of deep models, such as the manifold structure of class distribution.



%




\ifCLASSOPTIONcaptionsoff
  \newpage
\fi



%

\bibliographystyle{IEEEtran}
\bibliography{references}

\begin{thebibliography}{10}
\providecommand{\url}[1]{#1}
\csname url@samestyle\endcsname
\providecommand{\newblock}{\relax}
\providecommand{\bibinfo}[2]{#2}
\providecommand{\BIBentrySTDinterwordspacing}{\spaceskip=0pt\relax}
\providecommand{\BIBentryALTinterwordstretchfactor}{4}
\providecommand{\BIBentryALTinterwordspacing}{\spaceskip=\fontdimen2\font plus
\BIBentryALTinterwordstretchfactor\fontdimen3\font minus
  \fontdimen4\font\relax}
\providecommand{\BIBforeignlanguage}[2]{{%
\expandafter\ifx\csname l@#1\endcsname\relax
\typeout{** WARNING: IEEEtran.bst: No hyphenation pattern has been}%
\typeout{** loaded for the language `#1'. Using the pattern for}%
\typeout{** the default language instead.}%
\else
\language=\csname l@#1\endcsname
\fi
#2}}
\providecommand{\BIBdecl}{\relax}
\BIBdecl

\bibitem{b2}
A.~B. Hamida, A.~Benoit, P.~Lambert, and C.~B. Amar, ``3-d deep learning
  approach for remote sensing image classification,'' \emph{IEEE Transactions
  on Geoscience and Remote Sensing}, vol.~56, no.~8, pp. 4420--4434, 2018.

\bibitem{b7}
S.~Shabbir and M.~Ahmad, ``Hyperspectral image classification-traditional to
  deep models: A survey for future prospects,'' \emph{arXiv preprint arXiv:
  2101.06116}, 2021.

\bibitem{b4}
Z.~Gong, P.~Zhong, Y.~Yu, and et~al, ``A cnn with multiscale convolution and
  diversified metric for hyperspectral image classification,'' \emph{IEEE
  TGRS}, vol.~57, no.~6, pp. 3599--3618, 2019.

\bibitem{b14}
Y.~Yuan, C.~Wang, and Z.~Jiang, ``Proxy-based deep learning framework for
  spectral–spatial hyperspectral image classification: Efficient and
  robust,'' \emph{IEEE TGRS}, vol.~60, pp. 1--15, 2021.

\bibitem{b10}
N.~He, M.~E. Paoletti, J.~M. Haut, L.~Fang, S.~Li, A.~Plaza, and J.~Plaza,
  ``Feature extraction with multiscale covariance maps for hyperspectral image
  classification,'' \emph{IEEE TGRS}, vol.~57, no.~2, pp. 755--769, 2018.

\bibitem{b11}
H.~Lee and H.~Kwon, ``Going deeper with contextual cnn for hyperspectral image
  classification,'' \emph{IEEE TIP}, vol.~26, no.~10, pp. 4843--4855, 2017.

\bibitem{b13}
B.~Xi, J.~Li, Y.~Li, R.~Song, Y.~Xiao, Y.~Shi, and Q.~Du, ``Multi-direction
  networks with attentional spectral prior for hyperspectral image
  classification,'' \emph{IEEE TGRS}, vol.~60, pp. 1--15, 2022.

\bibitem{b12}
M.~E. Paoletti, J.~M. Haut, R.~Fernandez-Beltran, J.~Plaza, A.~J. Plaza, and
  F.~Pla, ``Deep pyramidal residual networks for spectral–spatial
  hyperspectral image classification,'' \emph{IEEE TGRS}, vol.~57, no.~2, pp.
  740--754, 2018.

\bibitem{b3}
Z.~Gong, P.~Zhong, and W.~Hu, ``Statistical loss and analysis for deep learning
  in hyperspectral image classification,'' \emph{IEEE TNNLS}, vol.~32, no.~1,
  pp. 322--333, 2021.

\bibitem{b1}
Z.~Gong, W.~Hu, X.~Du, P.~Zhong, and P.~Hu, ``Deep manifold embedding for
  hyperspectral image classification,'' \emph{IEEE Transactions on
  Cybernetics}, 2021.

\bibitem{b5}
H.~Huang, A.~Yu, and R.~He, ``Memory oriented transfer learning for
  semi-supervised image deraining,'' in \emph{IEEE CVPR}, 2021, pp. 7732--7741.

\bibitem{b8}
D.~Manolakis, E.~Truslow, M.~Pieper, and et~al., ``Detection algorithms in
  hyperspectral imaging systems: An overview of practical algorithms,''
  \emph{IEEE Signal Processing Magazine}, vol.~31, no.~1, pp. 24--33, 1992.

\bibitem{b9}
N.~M. Nasrabadi, ``Hyperspectral target detection: An overview of current and
  future challenges,'' \emph{IEEE Signal Processing Magazine}, vol.~31, no.~1,
  pp. 34--44, 2013.

\bibitem{b15}
Y.~Wen, K.~Zhang, Z.~Li, and Y.~Qiao, ``A discriminative feature learning
  approach for deep face recognition.''\hskip 1em plus 0.5em minus 0.4em\relax
  European Conference on Computer Vision, 2016, pp. 499--515.

\bibitem{b18}
``Pavia university data, accessed on may 21, 2022,'' https://www.
  ehu.eus/ccwintco/index.php/Hyperspectral\_Remote\_Sensing\_Scenes.

\bibitem{b16}
C.~Debes, A.~Merentitis, R.~heremans, J.~Hahn, N.~Frangiadakis, T.~van
  Kasteren, and et~al., ``Hyperspectral and lidar data fusion: Outcome of the
  2013 grss data fusion contest,'' \emph{IEEE Journal of Selected Topics in
  Applied Earth Observations and Remote Sensing}, vol.~7, no.~6, pp.
  2405--2418, 2014.

\bibitem{b19}
M.~E. Paoletti, J.~M. Haut, R.~Fernandez-Beltran, J.~Plaza, A.~J. Plaza, and
  F.~Pla, ``Deep pyramidal residual networks for spectral–spatial
  hyperspectral image classification,'' \emph{IEEE Transactions on Geoscience
  and Remote Sensing}, vol.~57, no.~2, pp. 740--754, 2018.

\bibitem{b20}
S.~K. Roy, G.~Krishna, S.~R. Dubey, and B.~B. Chaudhuri, ``Hybridsn: Exploring
  3-d-2-d cnn feature hierarchy for hyperspectral image classification,''
  \emph{IEEE Geoscience and Remote Sensing Letters}, vol.~17, no.~2, pp.
  277--281, 2019.

\bibitem{b21}
L.~Sun, G.~Zhao, Y.~Zheng, and Z.~Wu, ``Spectral-spatial feature tokenization
  transformer for hyperspectral image classification,'' \emph{IEEE Transactions
  on Geoscience and Remote Sensing}, 2022.

\end{thebibliography}




%




%


\vfill


\end{document}